\documentclass[runningheads]{llncs}
\usepackage{graphicx}
\usepackage{hyperref}
\usepackage{caption}
\usepackage{subcaption}

\usepackage{xcolor}

\begin{document}
\title{Selecting Optimal Trace Clustering Pipelines with AutoML}
%
%
\author{Sylvio Barbon Junior\inst{1}\orcidID{0000-0002-4988-0702} \and
Paolo Ceravolo\inst{2}\orcidID{0000-0002-4519-0173} \and
Ernesto Damiani\inst{3}\orcidID{0000-0002-9557-6496} \and
Gabriel Marques Tavares\inst{2}\orcidID{0000-0002-2601-8108}}
\authorrunning{Barbon et al.}
%
\institute{Londrina State University (UEL), Londrina, Brazil\\
\email{barbon@uel.br} \and
Universit\`a degli Studi di Milano (UNIMI), Milan, Italy\\
\email{\{paolo.ceravolo, gabriel.tavares\}@unimi.it} \and
Khalifa University (KUST), Abu Dhabi, UAE\\
\email{ernesto.damiani@kustar.ac.ae}}
\maketitle              
\begin{abstract}
Trace clustering has been extensively used to preprocess event logs. By grouping similar behavior, these techniques guide the identification of sub-logs, producing more understandable models and conformance analytics. Nevertheless, little attention has been posed to the relationship between event log properties and clustering quality. In this work, we propose an Automatic Machine Learning (AutoML) framework to recommend the most suitable pipeline for trace clustering given an event log, which encompasses the encoding method, clustering algorithm, and its hyperparameters. Our experiments were conducted using a thousand event logs, four encoding techniques, and three clustering methods. Results indicate that our framework sheds light on the trace clustering problem and can assist users in choosing the best pipeline considering their scenario.

\end{abstract}
\section{Introduction}

The execution of a business process leaves trails of the accomplished activities, performances achieved, and resources consumed. This information is stored in event logs which brace the history of the process. The executions generating the same sequence of activities are observed as the same trace by Process Mining (PM) algorithms that can group multiple executions in a single representation. Often, the variability of traces is however remarkable, and traces by themselves do not offer a helpful representation of the process. This variability causes problems for existing PM techniques. For instance, business processes with high trace variability generate spaghetti-like models, i.e., complex models with an enormous number of relations, often unreadable for the final user~\cite{Back2019}. 

Trace clustering techniques have been adopted to solve this issue by identifying sub-logs grouped by trace similarity. This way, by detecting groups with homogeneous behavior, process discovery techniques can be executed in these sub-logs, producing higher quality models, which are instead accessible for stakeholders \cite{10.1007/978-3-030-66498-5_21}. Trace clustering has also been studied in the context of explainability for PM \cite{DeKoninck2016}, and, more recently, adapted to incorporate expert knowledge \cite{DeKoninck2021}. However, selecting the appropriate clustering technique is not simple. Many transformation methods were presented, treating traces as vectors generated from bags of activities \cite{10.1007/978-3-540-78238-4_4}, edit distance~\cite{Bose2009} or dependency spaces~\cite{DELIAS2015203}, discriminant rules~\cite{1644726,10.1007/978-3-642-00328-8_11} or log footprints~\cite{DeKoninck2016}. The set of clustering algorithms applied is also ample, e.g., \textit{k}-means \cite{1644726}, hierarchical clustering \cite{Bose2009}, spectral clustering \cite{DELIAS2015203}, constrained clustering \cite{DeKoninck2021}, among others. Given this large set of options to setup a clustering pipeline, a non-expert user can likely feel overwhelmed.

Considering the challenge of designing pipelines for identifying the correct encoding method, clustering algorithm, and hyperparameters to be used for a specific log, we propose an AutoML framework based on  Meta-learning (MtL). Our framework recommends the trace clustering pipeline that best fits a specific event log. MtL is a learning process applied to meta-data representing other learning processes and has been used successfully to emulate expert's recommendations, maximize performance, and improve quality metrics~\cite{he2021automl}. In this work, the meta-data consist of a large set of event log features that are provided in input to the MtL workflow that outputs trace clustering pipelines described by an encoding technique, a clustering algorithm, and hyperparameters.  In our scenario, MtL learning serves as an AutoML approach as it suppresses the need for expert interaction to work properly. The relationship between event log features and quality of PM techniques has been already pointed out in the literature~\cite{10.1007/978-3-030-70650-0_11,barbon2021using}.  In this work, we introduce a general framework for studying this relationship for the trace clustering task using MtL. Moreover, we instantiate this framework to provide an example of its functionality. In particular, in our experiments we submit the method to a set of 1091 event logs described by 93 log features, four encoding techniques (\textit{one-hot}, \textit{position profiles}, \textit{bi-gram}, and \textit{tri-gram}), and three clustering algorithms (\textit{k}-means, \textit{dbscan}, and \textit{agglomerative}). Results show that our approach achieves 0.77 and 0.61 F1 scores for recommending encoding and clustering techniques, respectively. We also provide a comparison with two baseline performances, highlighting the improvement supported by the MtL strategy. Although, the same framework could be applied to other techniques to further investigate the domain.

The remainder of this paper is organized as follows. Section \ref{rw} gives a historical overview of trace clustering solutions, focusing on the employed transformation and clustering methods. Section \ref{ps} defines the problem and its configuration steps, while Section \ref{auto} presents our proposed framework to solve the trace clustering recommendation issue. Section \ref{es} presents the material used for experiments, the techniques, and quality metrics adopted. Section \ref{res} shows the results and raises a discussion around them. Section \ref{cc} concludes the paper.

\section{Related Work}\label{rw}

Trace clustering research is deeply connected to the variant analysis problem, that is, detecting groups of similar behavior within a single business process \cite{DeKoninck2016}. As stated by Koninck et al. \cite{DeKoninck2021}, clustering traces is partitioning an event log into groups of comparable traces such that each trace is assigned to a unique group, named cluster. Since its initial adoption, trace clustering has been proposed as an instrument to reduce variability. Discovering process models from clusters, for example, generally improves quality~\cite{10.1007/978-3-030-66498-5_21}. 
An early work in the area, presented by Greco et al. \cite{1644726}, uses a set of n-grams to encode a trace activity sequence, thus, transforming traces to feature vectors and input clustering techniques. Song et al. \cite{10.1007/978-3-642-00328-8_11} went further by defining multiple encoding procedures, named profiles, to represent traces as vectors. Furthermore, the authors call attention to the modularity between the profiling and clustering steps. Bose and van der Aalst \cite{Bose2009} represent traces as strings and apply edit distance to measure trace similarity. Delias et al. \cite{DELIAS2015203} proposed a measure to calculate trace distance based on dependency. Following a similar line, Appice and Malerba \cite{Appice2016} developed a co-training strategy to cluster traces based on multiple perspectives. The clusters are created using similarity measures based on multiple dimensions, namely activity, resource, sequence, and time difference. However, approaches based on instance-level similarity may be applicable only to particular domains depending on how the similarity is extracted. Thaler et al. \cite{thaler2015comparative} highlight that bags of activities may lose key information regarding the execution order. Delias et al. \cite{DELIAS2015203}, show that no single optimal similarity metric is applicable for all domains and applications. Zandkarimi et al. \cite{9230304} stated that trace clustering is a context-specific task. A better clarification of the problem is achieved by Koninck et al. \cite{DeKoninck2016}, which characterize the complexity of clustering with the assessment of the best event log splitting operations. Considering the plethora of available profiling techniques and clustering algorithms, we envision two main building blocks regulating the success of clustering techniques. The first regards the encoding method, converting the trace sequences into feature vectors or computing similarity metrics. The latter comprises the clustering techniques as given the algorithm's availability, one may not manage to choose a method. The approaches currently available in the literature are strictly attached to a specific combination of encoding and clustering algorithms; hence, they do not offer a means to study the relationship between the different steps that can compose a pipeline.

\section{Problem Statement}\label{ps}

As seen in Section \ref{rw}, past research has gathered heterogeneous approaches to the trace clustering problem. An expert may be able to assess business process characteristics and relate them to clustering approaches. However, given the plethora of configuration steps and parametrization, designing the appropriate trace clustering pipeline is a complex issue even for experts.

We identified in the literature three configuration steps that highly affect the clustering results: (i) trace encoding, (ii) clustering algorithm, and (iii) hyperparameters regulating the clustering algorithm. The choice of each step is critical since slight changes deeply affect the clustering results, hindering the accessibility of solutions for non-expert users. Regarding trace encoding, Barbon et al. \cite{10.1007/978-3-030-70650-0_11} stated that a well-performing encoding method improves a wide range of posterior analyses without the need of tuning them. In PM applications, encoding aims at transforming traces into mathematical representations, most frequently vectors, which map process instances into a feature space. The authors also showed that there is no best encoding method for every scenario in the anomaly detection task, that is, different event logs are encoded better, considering several quality criteria, by different encoding techniques. A similar conclusion is achieved by Thaler et al. \cite{thaler2015comparative} when analyzing clustering algorithms applied to PM. The authors stated that some techniques are suitable for particular scenarios, reinforcing the argument that process characteristics may guide the decision of the appropriate clustering technique. Besides, different from supervised approaches, unsupervised learning performance is severely affected by small changes in hyperparameters, depending heavily on user-domain knowledge~\cite{7460951}. This implies the solutions proposed today are far from optimal as they are attached to a unique set of encoding and clustering algorithms.

\section{AutoML as a Solution for Trace Clustering}\label{auto}

\begin{figure}[ht!]
    \centering
    \includegraphics[width=0.65\textwidth]{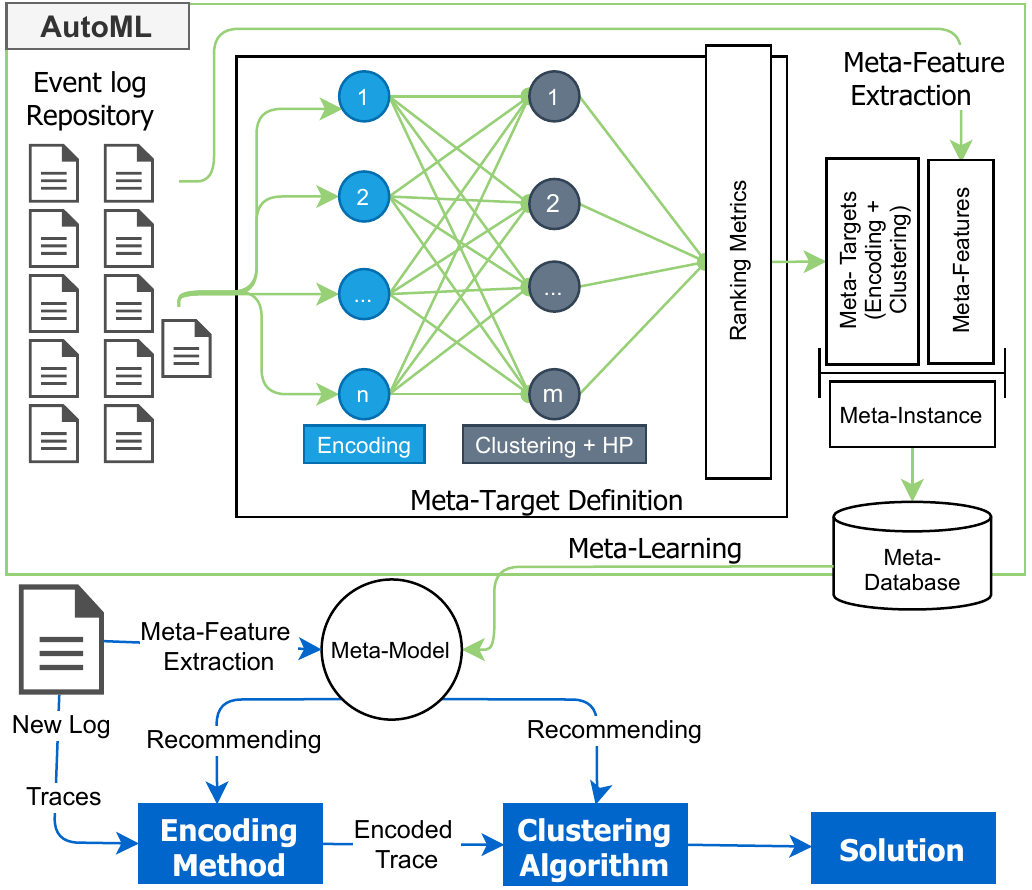}
    \caption{Overview of AutoML proposal for Trace Clustering.}
    \label{overview}
\end{figure}

Trace clustering solutions must be able to adapt according to domain characteristics. We then propose a framework grounded in AutoML capable to deliver suitable recommendations according to different business process behaviors. The main goal of our approach is recommending a tuple $\langle$\textit{encoding}, \textit{clustering}, \textit{hyperparameters}$\rangle$ that maximizes quality metrics for the trace clustering problem. Fig. \ref{overview} shows the overview of building blocks controlling the framework. First, an event log repository is created to represent different business scenarios. The \textit{Meta-feature Extraction} step mines features for each event log in the repository, creating \textit{meta-features} according to MtL terminology. The description quality of the \textit{meta-features} is an important constraint affecting the performance of the complete pipeline. Moreover, the \textit{Meta-target Definition} defines a set of encoding and clustering (coupled with its hyperparameter) techniques that are assessed by quality metrics and ranked according to a ranking function. Then, the \textit{Meta-database} combines the \textit{meta-features} and \textit{meta-targets} defined in previous steps, creating a data set populated by \textit{meta-instances}. Using the \textit{meta-database}, the \textit{Meta-learning} step induces a \textit{Meta-model} that is, then, used to recommend a tuple $\langle$\textit{encoding}, \textit{clustering}, \textit{hyperparameters}$\rangle$ for a given event log considering its \textit{meta-features}. It is worth mentioning that multi-output machine learning modeling for the meta-model can bring important achievements in terms of performance considering the interrelations between each step of the pipeline. In Fig. \ref{overview}, green arrows indicate the steps that are used for the creation and training of the framework, while blue arrows represent a production environment where one assesses the meta-model for recommending.

Given the adaptable setup of our framework, one can implement it using a different set \textit{meta-features} and \textit{meta-targets}. The automatic aspect of this approach provides the user with recommendations based on event log behavior considering the possible options among the configurable steps. Moreover, other aspects are adaptable, such as the adopted quality metrics and the ranking function. Nonetheless, we note that the robustness of the approach depends on the AutoML structure, which must be maintained when the framework is instantiated in real scenarios.

\section{Experimental Setup}\label{es}

In this section, we expose the details regarding the experiments implemented to study a possible instance of our AutoML framework. This is obtained by choosing specific techniques for generating the \textit{meta-features} (event log features) and the \textit{meta-targets} (trace encoding and trace clustering with hyperparameters). The implementation is available for replication purposes\footnote{\url{https://github.com/gbrltv/process\_meta\_learning}}.

\subsection{Event logs and featurization}

MtL benefits from using a large set of instances in the meta-database. Hence, we are aiming at a heterogeneous set of business process logs, representing different scenarios and behaviors. For that, we rely on the set of logs proposed by Barbon et al. \cite{barbon2021using}. These event logs were grouped to represent a plethora of business behaviors, mapping the relationship between process characteristics and quality metrics. This set contains both real and synthetic event logs. Regarding real-life data, there are six logs from past Business Process Intelligence Challenges (BPIC)\footnote{\url{https://www.tf-pm.org/resources/logs}}, the environmental permit\footnote{\url{https://doi.org/10.4121/uuid:26aba40d-8b2d-435b-b5af-6d4bfbd7a270}}, helpdesk\footnote{\url{https://doi.org/10.17632/39bp3vv62t.1}} and sepsis\footnote{\url{https://doi.org/10.4121/uuid:915d2bfb-7e84-49ad-a286-dc35f063a460}} logs. For synthetic data, the authors adopted 192 logs from the Process Discovery Contest (PDC) 2020\footnote{\url{https://www.tf-pm.org/competitions-awards/discovery-contest}}, an annual event organized to evaluate the efficiency of process discovery algorithms. The PDC logs are complex given the nature of employed behaviors, such as dependent tasks, loops, invisible and duplicate tasks, and noise. The next group of synthetic data contains 750 logs proposed in the context of online PM \cite{9124702}. These logs are built to depict process drifts, i.e., behavior change during the business process execution. For that, a model was created and perturbed by 16 change patterns, representing different changes from the original model. Moreover, the logs contain four drift types, five noise percentages, and three trace lengths. The final group of synthetic event logs was proposed for the evaluation of trace encoding techniques \cite{10.1007/978-3-030-70650-0_11}. This set contains 140 logs generated from five process models, six anomaly types, and four frequency percentages.

The performance of the meta-model is directly dependent on the quality of the meta-features. Thus, the group of meta-features extracted from event logs must correctly capture the process behavior and describe it from complementary perspectives. As proposing log descriptors is out of the scope of this work, we adopted the featurization introduced in \cite{barbon2021using}. The authors presented a group of features that capture several layers of business processes, i.e., activity, trace, and log. Regarding activity-level features, the group is subdivided into: all activities, start activities, and end activities. 12 features are extracted for each group, they are the number of activities, minimum, maximum, mean, median, standard deviation, variance, the 25th and 75th percentile of data, interquartile range, skewness, and kurtosis coefficients. To capture behavior at the trace-level, the authors propose features for trace lengths and trace variants. The former group contains 29 attributes: minimum, maximum, mean, median, mode, standard deviation, variance, the 25th and 75th percentile of data, interquartile range, geometric mean and standard variation, harmonic mean, coefficient of variation, entropy, and a histogram of 10 bins along with its skewness and kurtosis coefficients. Trace variants are captured by 11 descriptors: mean number of traces per variant, standard variation, skewness coefficient, kurtosis coefficient, the ratio of the most common variant to the number of traces, and ratios of the top 1\%, 5\%, 10\%, 20\%, 50\% and 75\% variants to the total number of traces. Log-level behavior is captured by: number of traces, unique traces, and their ratio, and number of events. Finally, to describe log complexity, entropy-based measures have been adopted recently in PM literature \cite{Back2019}. The entropy metrics proposed in \cite{Back2019} aim at the discretization between logs that are better mined by declarative or imperative algorithms. Hence, such metrics capture log structuredness and variability. The 14 entropy features we adopt are: trace, prefix, \textit{k}-block difference and ratio (applied with \textit{k} values of 1, 3 and 5), global block, \textit{k}-nearest neighbor (applied with \textit{k} values of 3, 5, and 7), Lempel-Ziv, and Kozachenko-Leonenko. Considering all groups, 93 meta-features were used to extract log characteristics.

\subsection{Trace encoding techniques}

Many PM techniques rely on encoding to transform event log-specific representations to other formats \cite{1316839,10.1007/978-3-319-23063-4_21,Polato2018,10.1007/978-3-319-98648-7_18}. The transformation usually applies at the trace-level, that is, converting the sequence of activities respective to a unique trace into a feature vector. In \cite{10.1007/978-3-030-70650-0_11}, the authors compared 10 different encoding techniques through the lens of quality metrics measuring data dispersity, representativeness, and compactedness. These encoding methods were inspired by three different families: PM native, word and graph embeddings. A classification task for anomaly detection was also employed to measure encoding quality. As pointed out by the authors, there is no encoding that excels in all tasks and perspectives concomitantly. For instance, graph embeddings outperform the others in the classification task and representation quality. However, these encoding methods are costly and usually sparse, meaning that there are better encoding techniques considering space and time complexity. The trace clustering literature has already experimented with several types of encoding methods. In \cite{1644726} and \cite{10.1007/978-3-642-00328-8_11}, the authors adopt the one-hot encoding technique to transform traces before the clustering step. In \cite{Bose2009}, the authors employ edit distance to compute the trace distance preceding the clustering. Koninck et al. \cite{DeKoninck2016} used log footprints, i.e., control-flow relations depicting activity sequences. In \cite{8786061}, the authors apply activity profiles, bi-gram and tri-gram as methods for trace encoding. Nonetheless, Leoni et al. \cite{DELEONI2016235} pointed that no trace similarity measure is general enough to be applicable in all scenarios.

In this work, we adopt four encoding techniques that were frequently applied in the context of trace clustering. The first one is one-hot encoding. This technique encodes activities as categorical dimensions, creating a feature vector of binary values for each trace, based on the occurrence of activities in a trace. Next, we adopt n-grams, a common technique used in text mining applications. This encoding maps groups of activities of size \textit{n} into a feature vector, accounting for their occurrence or not. More specifically, we apply bi-gram and tri-gram. Finally, we applied position profiles \cite{10.1007/978-3-319-65015-9_4}, an approach that relates activity frequency and position. A log profile is created by computing the activity appearances in each trace position and its respective frequency. It follows that a trace is encoded considering the frequency of its activities in their positions according to the log profile.

\subsection{Trace clustering algorithms}

We selected three clustering techniques commonly applied in data mining and trace clustering literature. These techniques are grounded in different heuristics, that is, each algorithm approaches the clustering problem from a unique perspective. With this, we aim at evaluating if a particular clustering structure outperforms the others. 

First, we adopt the Density-based Spatial Clustering of Applications with Noise (\textit{dbscan}) algorithm \cite{Ester:1996:DAD:3001460.3001507}. The \textit{dbscan} method guides its clustering based on the density of the feature space, hence, instances in high-density regions form a cluster while instances sitting at low-density regions are regarded as outliers. The main hyperparameter affecting the clustering results is \textit{eps}, which regulates the maximum distance between two points for them to be considered of the same neighborhood. We explore different configurations of the \textit{eps} hyperparameter to evaluate its impact and to recommend the best configuration in the meta-model step. For that, we apply the following \textit{eps} values: 0.001, 0.005, 0.01, 0.05, 0.1, 0.5, 1, 5, 10, 50. Moreover, we adopt \textit{k}-means \cite{macqueen1967some}, a clustering technique that randomly selects centroids, which are the initial cluster points, and works by iteratively optimizing the centroid positions. The optimization stops either when centroid positions are stable or when the maximum number of repetitions is achieved. The \textit{k}-means technique requires the expected number of clusters (\textit{k}) from a given data set as a hyperparameter. We set \textit{k} to these values: 2, 3, 4, 5, 6, 7, 8, 9, 10. Finally, the last technique is \textit{agglomerative} clustering \cite{Ward1963}, a type of hierarchical clustering with a bottom-up approach. The algorithm starts by considering each point as a cluster. Further, it merges the clusters as the hierarchy moves up, creating a tree-like structure depicting the cluster levels and merges. Cluster pairs are merged given a linkage distance, two clusters that minimally impact the linkage distance are merged recursively. As with \textit{k}-means, \textit{agglomerative} clustering requires the number of clusters to be found, we then adopted the same range of values for the \textit{k} parameter.

\subsection{Ranking metrics}\label{sec:rm}

To complete the creation of a meta-database, meta-targets must be defined for each meta-instance. This way, a ranking strategy is required to compare both encoding and clustering techniques. Hence, the technique sitting at the top of the ranking strategy is the one recommended for a meta-instance, i.e., it is defined as the meta-target. As pointed out in the literature \cite{10.1007/978-3-030-70650-0_11,DELEONI2016235}, there is no unique solution for a problem that outperforms the others from all perspectives. Considering this hypothesis, we propose three complementary metrics to evaluate trace clustering solutions, this way, capturing different degrees of performance. Moreover, a user applying a trace clustering solution may expect to evaluate the results from several perspectives. Here, we support such a user by assessing clustering quality from a set of criteria.

Silhouette coefficient, the first metric we propose to measure performance, is based on the traditional clustering literature, commonly applied in data mining domains \cite{ROUSSEEUW198753}. The Silhouette score is computed at the cluster level to capture its tightness and separation, judging instances that fit their cluster or are in between different clusters. The scores of a group of clusters can be combined to assess the relative quality of the clustering technique. Equation \ref{eq:sil} demonstrates how the Silhouette coefficient ($s$) is obtained for a single sample considering the mean intra-cluster distance ($a$) and the mean nearest-cluster distance ($b$). The average of the Silhouette score for all samples is the final coefficient for one clustering space, that is, the average of Equation \ref{eq:sil} for all samples in the feature space. The Silhouette coefficient domain is $[-1, 1]$, where $-1$ is the worst value, $0$ indicates overlapping clusters and $1$ is the best value.

\begin{equation}\label{eq:sil}
    s = \frac{b - a}{max(a, b)}
\end{equation}

To complement this evaluation with a PM-inspired metric, we propose to measure the quality of clusters concerning trace variants. This way, by computing the trace variant frequency in each cluster, we can evaluate if the solution provides a clear separation of variants in the feature space. For that, we compute the unique traces in a cluster, and by a weighted mean, the Variant score is reached. Consider $C_i$ the cluster of index $i$, $C$ the group of all clusters, $\#variants$ the number of unique traces found in cluster $C_i$ and $\#traces$ the total number of traces in the event log, Equation \ref{eq:var} depicts the Variant score calculation, $0$ is the optimal value.

\begin{equation}\label{eq:var}
    v = \frac{\sum_{C_i \in C}^{} \#variants - 1}{\#traces}
\end{equation}

As resource consumption is an important aspect in organizations, we also consider the computational time ($t$) of clustering as a metric to assess its quality. The lower the $t$ metric for a particular solution, the better it is ranked in comparison to others. Given this set of metrics, i.e. $s$ for silhouette coefficient, $v$ for variant score, and $t$ for computational time, a \textit{meta-target} $\langle$\textit{encoding}, \textit{clustering}, \textit{hyperparameters}$\rangle$ has to successfully balance between all metrics to be considered a good set. This way, our approach rewards techniques that excel in the three metrics, such as ignoring one or more may lead to lack of tightness, improper variant identification, and high resource consumption. Hence, we propose a ranking strategy ($R$) that combines all dimensions. Table \ref{tab:ranking} presents an example of the ranking strategy we propose. For each pair of encoding techniques and clustering algorithms, we apply it for a given event log ($L$) and measure the quality metrics ($s$, $v$, $t$). Following, a rank is built for each metric ($R_s$, $R_v$, $R_t$), i.e, comparing the pairs of encodings and clustering in each dimension. Finally, a rank ($R$) is computed by the average of the metrics ranks. For example, considering the pairs $\langle E_1, C_1 \rangle$, $\langle E_2, C_2 \rangle$ and $\langle E_3, C_3 \rangle$, their respective final ranks are $2$, $1.67$ and $2.33$. The solution chosen as the meta-target is the one that minimizes the $R$ function, in this example the pair $\langle E_2$, $C_2 \rangle$.

\begin{table}[!ht]
\centering
\begin{tabular}{llllllllll}
\hline
Log & Encoding & Clustering & $s$ & $v$ & $t$ & $R_s$ & $R_v$ & $R_t$ & $R$\\
\hline
$L$ & $E_1$ & $C_1$ & 0.9 & 0.5 & 50 & 1 & 2 & 3 & 2\\
$L$ & $E_2$ & $C_2$ & 0.3 & 0 & 10 & 3 & 1 & 1 & 1.67\\
$L$ & $E_3$ & $C_3$ & 0.8 & 0.7 & 15 & 2 & 3 & 2 & 2.33\\
\hline
\end{tabular}
\caption{Example of ranking encoding and clustering pairs. The final rank function is the average rank of each quality dimension.}
\label{tab:ranking}
\end{table}

\subsection{Meta-model}

Before introducing the meta-learner used to create the meta-model, we first provide some particular details about the type of problem faced in our AutoML framework. As presented in Section \ref{sec:rm}, we need indeed to suggest both the best trace clustering algorithm and the best trace encoding technique. This implies that given a meta-instance the system recommends the tuple that achieves the maximum performance for the combined metrics. Most research in supervised learning proposes algorithms for \textit{single-label} problems, where instances are associated with a single label $\lambda$ from a set of disjoint labels $L$. However, in the proposed setup, we are facing a \textit{multi-output} problem, where a set of labels $Y \subseteq L$ is associated with a single instance \cite{Tsoumakas2010}. Following the taxonomy proposed by Tsoumakas et al. \cite{Tsoumakas2010}, we adopt a problem transformation approach, which converts the data into a format that can be used in conjunction with traditional techniques. More specifically, we employed the Binary Relevance (BR) transformation approach. BR works by transforming the original data set into $q$ data sets $D_{\lambda_j}$, where $j = [1, ..., q]$ contains all instances of the original data that are labeled according to the existence or not of $\lambda_j$. Thus, BR learns $q$ binary classifiers, one for each label $L$. Given a new instance, BR provides the union of the labels $\lambda_j$ predicted by the $q$ classifiers.

Regarding the meta-learner, we applied the Random Forest (RF) algorithm \cite{breiman2001random} due to its robustness, being less prone to overfitting. RF creates a collection of decision trees with a bagging technique, i.e., randomly selecting features for each tree. This way, our meta-model combines the RF with the BR approach. Moreover, we applied a simple hyperparameter tuning technique to improve performance in the recommendation task. For that, we divided the meta-database into three sets: train, validation, and test, respectively containing 80\%, 10\%, and 10\% of the total number of meta-instances. The grid search strategy was used for tuning. This method exhaustively evaluates all combinations of chosen hyperparameters and uses cross-validation splitting to capture an average performance. The results reported in Section \ref{res} were extracted when applying the tuned meta-model to the test set. The hyperparameters tuned were: (i) the number of trees composing the forest, (ii) the criterion measuring split quality, (iii) the minimum required number of samples for a node split, (iv) the minimum number of samples required to be a leaf node, and (v) the number of considered features for a split.

\section{Results and Discussion}\label{res}

In this section, we present and discuss the main experimental results regarding the proposed strategy to recommend a trace clustering pipeline based on AutoML. We started by exploring the meta-database composition by observing the encoding techniques and clustering algorithms chosen by their performance and balancing. Next, an overall analysis, including the comparison of the proposed strategy with the baselines (\textit{random} and \textit{majority}), is introduced, while a detailed assessment of meta-features is presented in the last part.

\subsection{Meta-Learning exploratory analysis}

\begin{figure*}[!ht]
    \centering
    \begin{subfigure}[b]{0.26\textwidth}    
       \includegraphics[width=1\textwidth]{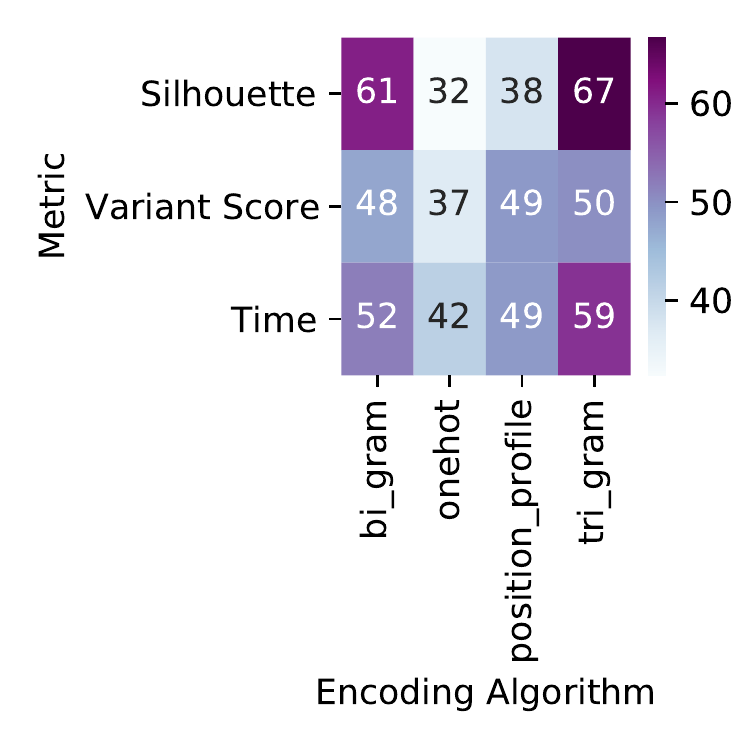}
    \caption{Encoding Ranking}
    \label{enc_rank}
    \end{subfigure}
    \begin{subfigure}[b]{0.73\textwidth}    
       \includegraphics[width=1\textwidth]{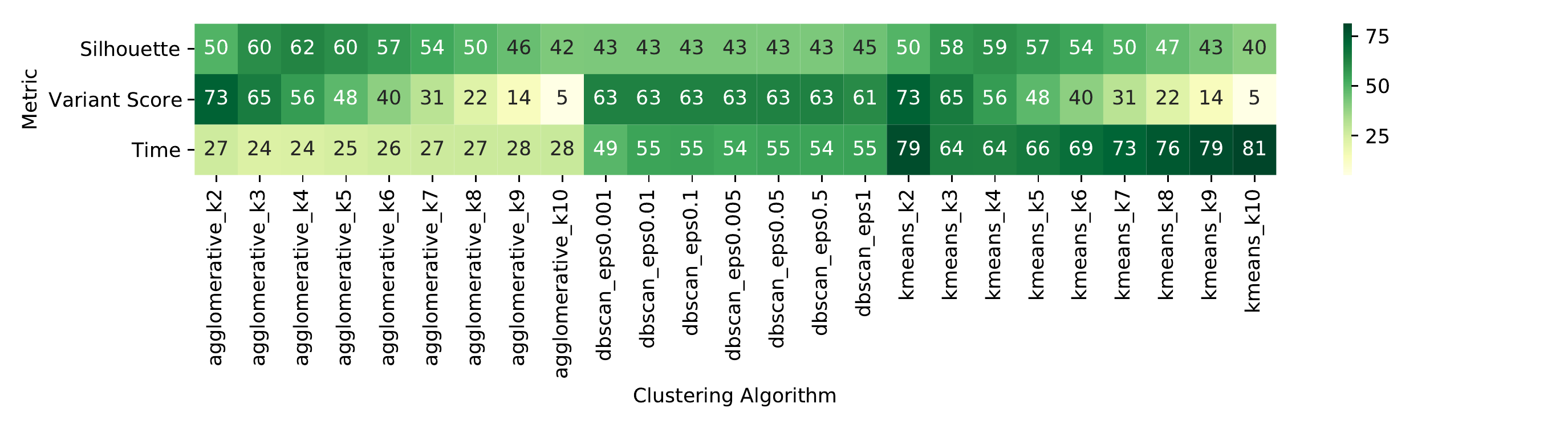}
    \caption{Clustering Ranking}
    \label{clus_rank}
    \end{subfigure}
    \caption{Ranking of encoding (a) and clustering (b) to identify the meta-target. Color variation represents the variation of ranking position.}
    \label{heatmap}
\end{figure*}

The results, considering all algorithms for setting the meta-database, including the metrics used for ranking the meta-targets, are presented in Fig. \ref{heatmap}.  The heat-map plots show the ranking of the metrics $s$, $v$, and $t$ for encoding (Fig.~\ref{enc_rank}) and clustering (Fig.~\ref{clus_rank}) used to sort and identify promising algorithms as meta-targets. Each ranking varies from 1 to 81, in which 1 is the best-ranked algorithm for a given metric.

Observing the encoding techniques (Fig.~\ref{enc_rank}), it is possible to identify a large discrepancy between them when evaluated by Silhouette, revealing the superiority of \textit{one-hot} and \textit{position profile} algorithms. Variant score and Time do not present a prominent variation such as Silhouette, leading to closer ranking positions. Based on these results, it is possible to support the hypothesis of the ``no free lunch theorem" due to the ranking balance since there is no best technique for all quality criteria concurrently. However, when observing the clustering algorithms (Fig.~\ref{clus_rank}), it is possible to note a balance regarding Silhouette, whereas Variant score and Time reveal discrepancies. The first one, Variant score, exposes the importance of hyperparameter definition since \textit{agglomerative} and \textit{k}-means ranged throughout the rankings, when changing their hyperparameter \textit{k}. Moreover, the Time metric delivered an important perspective, in which each clustering algorithm is recognizable regardless of its hyperparameters. In particular, \textit{agglomerative} and \textit{dbscan} were superior to \textit{k}-means. This superiority led to no usage of \textit{k}-means as a clustering meta-target.

\begin{figure}[ht!]
    \centering
    \includegraphics[width=0.65\textwidth]{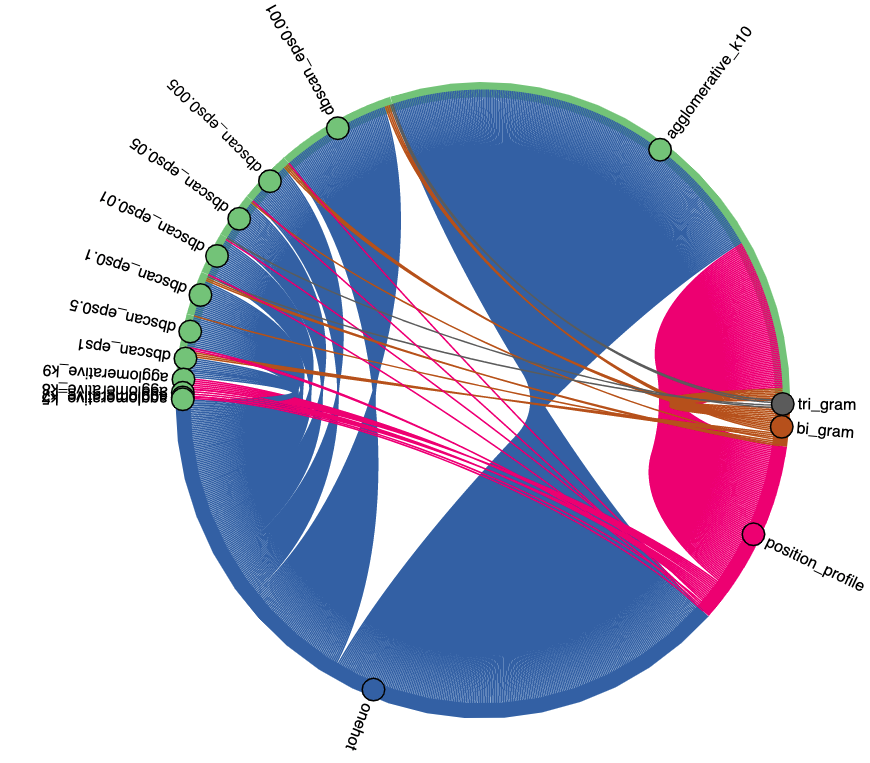}
    \caption{The combinations of encoding techniques and clustering algorithms are links, which represents a meta-instance that best fit linked algorithms colored by encoding.}
    \label{fig:chord}
\end{figure}

The meta-database was built using the combination of the top ranked algorithms for each meta-instance (event logs). This combination leads to an imbalanced multi-output dataset, which was handled to support the induction of the meta-model. This imbalanced scenario can be seen in Fig. \ref{fig:chord}, where combinations such as \textit{one-hot} enconding with \textit{agglomerative} clustering using 10 as $k$ value  ($onehot\_agglomerative\_k10$) represented 469 meta-instances. The second most frequent combination was \textit{position profile} with \textit{agglomerative} clustering using 10 as $k$ value ($position\_profile\_agglomerative\_k10$), reaching 171 meta-targets. The third was \textit{one-hot} using \textit{dbscan} adopting a \textit{eps} equals 0.001 ($onehot\_dbscan\_eps0.001$) in 125 meta-instances. Fig. \ref{fig:chord} represents in blue the \textit{one-hot} combinations, in pink the \textit{position profile}, \textit{bi-gram} is brown and \textit{tri-gram} gray. The domination of \textit{one-hot}, followed by position profile and bi-gram is evident. \textit{Tri-gram} was the best one, combined with \textit{dbscan}, only with four meta-instances. 

\begin{figure}[ht!]
    \centering
    \includegraphics[width=0.65\textwidth]{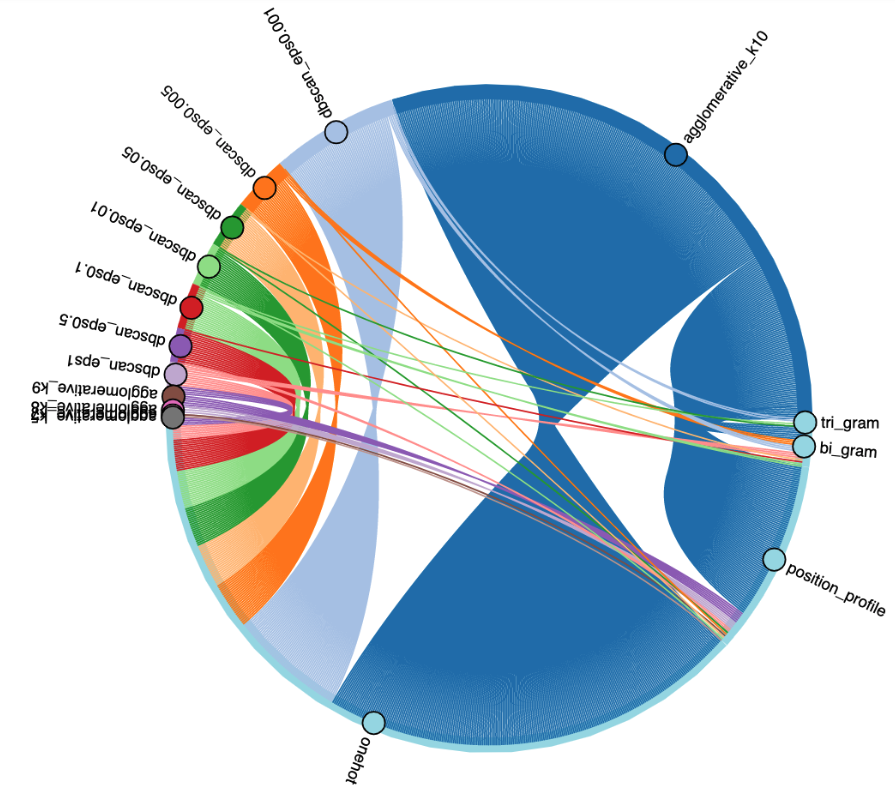}
    \caption{The combinations of encoding and clustering algorithms are links, which represents a meta-instance that best fit linked algorithms colored by clustering.}
    \label{fig:chord2}
\end{figure}

When evaluating from an encoding perspective (Fig. \ref{fig:chord2}), we observe a balance between \textit{dbscan} with a wide range of \textit{eps} and \textit{agglomerative} using $k$ equals 10. Different values of $k$ for \textit{agglomerative} did not meet many meta-instances. Conversely, \textit{dbscan} demonstrate the necessity of hyperparameter adjustments since different values of \textit{eps} could match particular meta-instances.

The imbalance issue was addressed by removing the minority classes combinations, that is, pairs of encoding techniques and clustering algorithms that appear as a meta-target for less than five meta-instances.
The final meta-database was composed of 1036 samples, with fifteen different combinations of \textit{one-hot}, \textit{position profile}, and \textit{bi-gram} with \textit{agglomerative} (\textit{k} in \{8, 9, 10\}) and \textit{dbscan} (\textit{eps} in \{0.001, 0.005, 0.05, 0.01, 0.1, 0.5, 1\}).

\subsection{Meta-Model performance}

Using RF as our meta-model built over the meta-database, we analyzed the performance for both encoding and clustering algorithm recommendations. It is worth mentioning that our problem was modeled as a multi-output problem, addressing encoding and clustering at once, taking advantage of possible inter-correlations between both steps.

\begin{figure}[ht!]
    \centering
    \includegraphics[width=0.6\textwidth]{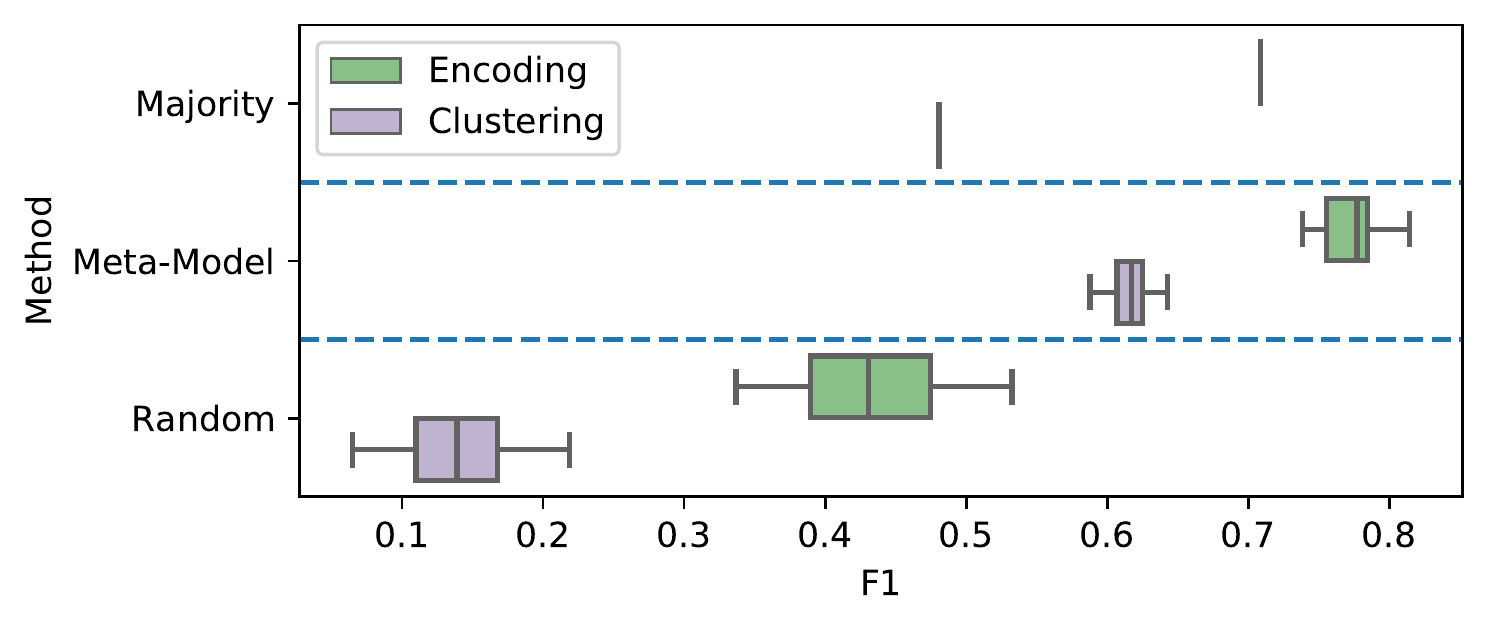}
    \caption{Performance of the AutoML framework to recommend the encoding technique and clustering algorithm in terms of accuracy and F1.}
    \label{fig:performance}
\end{figure}

Our proposal obtained an F1 of $0.77$ ($\pm 0.01$) when recommending the encoding technique and an F1 of $0.61$ ($\pm 0.01$) for clustering algorithm recommendation. To bring insights on the performance achieved, we compared the results with the majority classes (\textit{one-hot} as encoding technique and $agglomerative\_k10$ as clustering algorithm) and with a random selection, seen in Fig.~\ref{fig:performance}. The majority baseline for encoding obtained an F1 of $0.70$ ($\pm 0.00$). The random baseline for encoding achieved $0.41$ ($\pm 0.04$) of F1. Considering clustering, the majority obtained $0.48$ ($\pm 0.00$) of F1 and random selection reached $0.14$ ($\pm 0.04$), respectively. Regarding the mean predictive performance in terms of F1, for the whole trace clustering pipeline, our proposed AutoML approach obtained $0.69$ ($\pm 0.07$). The results were superior to the majority and random baselines, which achieved $0.59$ ($\pm 0.11$) and $0.28$ ($\pm 0.15$), respectively. Note that the majority results are boosted by the imbalanced scenario, for balanced meta-databases, the tendency is to underperform.

\subsection{Meta-features relevance}

\begin{figure}[ht!]
    \centering
    \includegraphics[width=0.65\textwidth]{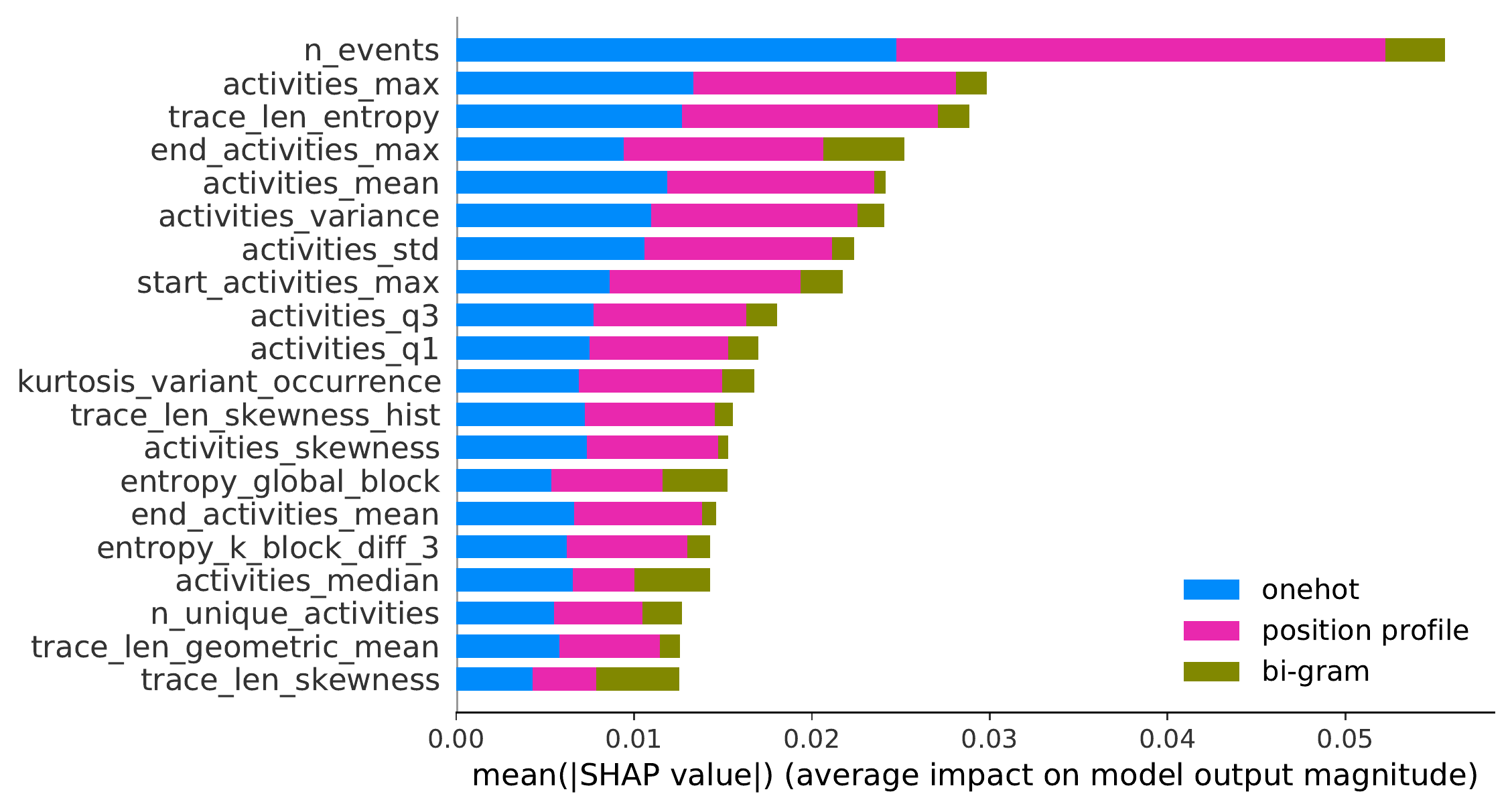}
    \caption{The relative importance for each feature, obtained by taking the average absolute value of the SHAP values when recommending encoding algorithms.}
    \label{fig:shap_enc}
\end{figure}

\begin{figure}[ht!]
    \centering
    \includegraphics[width=0.65\textwidth]{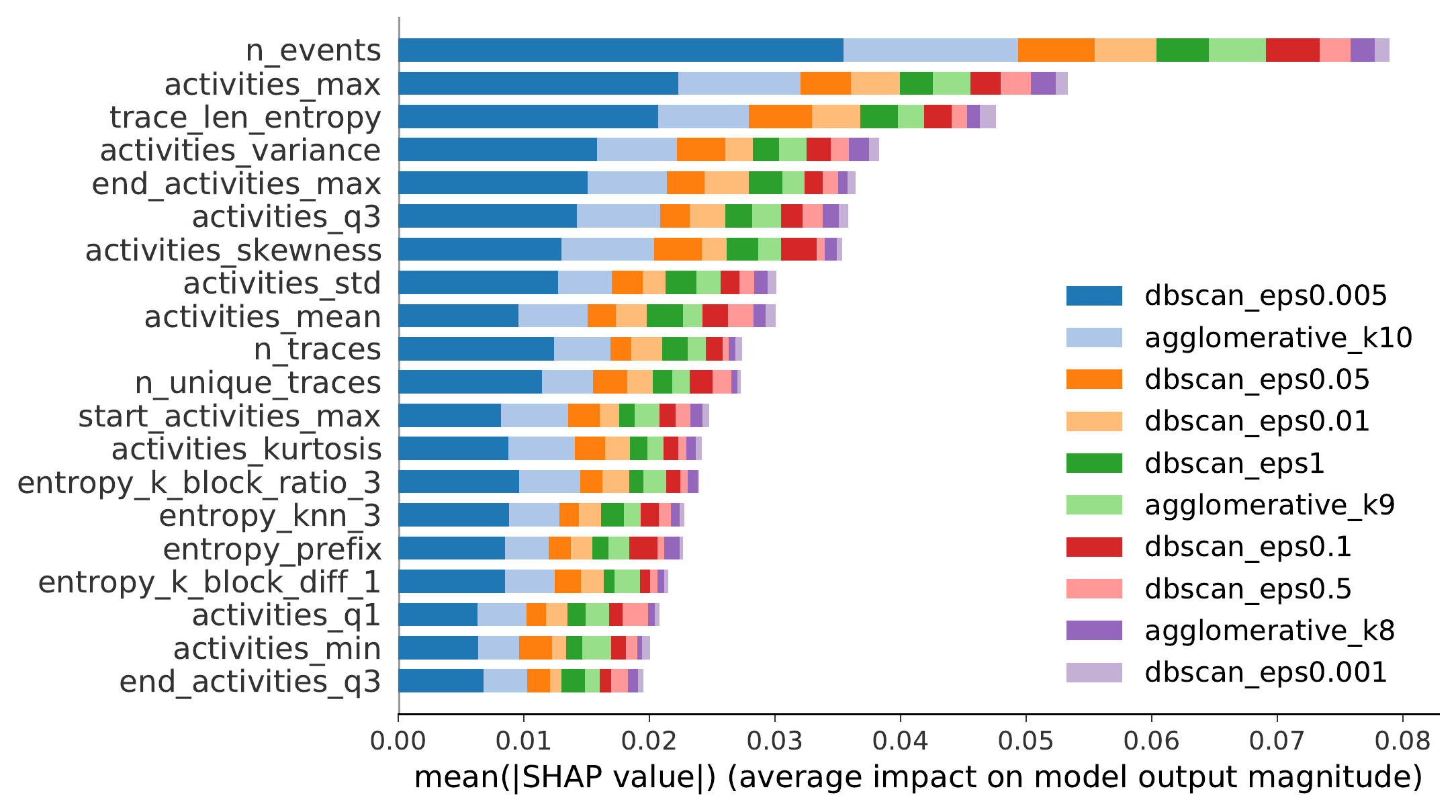}
    \caption{The relative importance for each feature, obtained by taking the average absolute value of the SHAP values when recommending clustering algorithms.}
    \label{fig:shap_clus}
\end{figure}

We interpreted the outputs of our meta-model to predict encoding (Fig.~\ref{fig:shap_enc}) and clustering (Fig.~\ref{fig:shap_clus}) by taking the average absolute value of the Shapley Additive Explanation (SHAP) values. The higher relative importance for predicting encoding algorithms were obtained by the number of events ($n\_events$), the maximum number of activities ($activities\_max$), and the entropy of trace length ($trace\_len\_entropy$). Similarly, the top three meta-features in terms of importance were $n\_events$, $activities\_max$ and $trace\_len\_entropy$ when predicting clustering. It is important to mention that $n\_events$ represented more than half of the importance among all meta-features for both encoding and clustering. The $n\_events$ meta-feature indirectly indicates log complexity as the higher the number of events, the more heterogeneous behavior might appear, even more when considering that many of the logs come from complex models and include anomalies. Thus, $n\_events$ becomes an important discriminator for encoding and clustering performances. These results highlight relevant directions for future research in feature extraction in PM.

\section{Conclusion}\label{cc}

In this paper, we proposed an AutoML framework to recommend the best pipeline for trace clustering based on a specific event log. For that, we extract meta-features to describe event logs and matched them with the best clustering pipeline by assessing three complementary metrics (Silhouette, Variant score, and Time). The framework recommends a tuple $\langle$\textit{encoding}, \textit{clustering}, \textit{hyperparameters}$\rangle$, making trace clustering solutions accessible for non-expert users. Results have shown that the framework outperforms baseline approaches. We have also provided a discussion about meta-feature influence in the decision process using SHAP values. In future research, we aim to extend the experimental evaluation to gather further insights into the relationship between trace clustering quality and event log behavior.

\bibliographystyle{splncs04}
\bibliography{bibliography}

\end{document}